\documentclass[sigconf]{aamas} 

\usepackage[colorinlistoftodos,textsize=tiny]{todonotes}

\usepackage{balance} % for balancing columns on the final page

\usepackage{listings}

\colorlet{punct}{red!60!black}
\definecolor{background}{HTML}{EEEEEE}
\definecolor{delim}{RGB}{20,105,176}
\colorlet{numb}{magenta!60!black}

\usepackage{lstcoq}
\usepackage{booktabs}
\usepackage{array}

\usepackage{lipsum}

\usepackage{tikz}
\usetikzlibrary{arrows.meta,positioning,shapes.misc, shapes.geometric, decorations.pathreplacing, fit, backgrounds,shapes.multipart, calc, decorations.pathmorphing} 
\usepackage{tikz-cd}

\usepackage{subfig}
\usepackage{soul}

\usepackage[utf8]{inputenc}
\usepackage[T1]{fontenc}
\usepackage{url}
\usepackage{amsfonts}
\usepackage{nicefrac}
\usepackage{microtype}
\usepackage{xcolor}
\usepackage{graphicx}
\usepackage{pdfpages}
\usepackage{orcidlink}

\usepackage{cleveref}

\usepackage[most]{tcolorbox}

\newcommand{\jetbrains}{JetBrains}

\setcopyright{ifaamas}
\acmConference[AAMAS '26]{Proc.\@ of the 25th International Conference
on Autonomous Agents and Multiagent Systems (AAMAS 2026)}{May 25 -- 29, 2026}
{Paphos, Cyprus}{C.~Amato, L.~Dennis, V.~Mascardi, J.~Thangarajah (eds.)}
\copyrightyear{2026}
\acmYear{2026}
\acmDOI{}
\acmPrice{}
\acmISBN{}

\acmSubmissionID{676}

\title[RocqStar]{RocqStar: Leveraging Similarity-driven Retrieval and Agentic Systems for Rocq generation}

\author{Andrei Kozyrev \orcidlink{0009-0004-3185-9368} }
\authornote{Authors contributed equally to this work.}
\affiliation{
  \institution{{\jetbrains} Research}
  \country{Germany}
}

\author{Nikita Khramov \orcidlink{0009-0004-3968-4443} }
\authornotemark[1]
\affiliation{%
  \institution{{\jetbrains} Research}
  \country{Germany}
}

\author{Gleb Solovev \orcidlink{0009-0004-1116-7743}}
\affiliation{%
  \institution{{\jetbrains} Research}
  \country{Germany}
}

\author{Anton Podkopaev \orcidlink{0000-0002-9448-6587}}
\affiliation{%
  \institution{{\jetbrains} Research}
  \country{the Netherlands}
}
\affiliation{%
  \institution{Constructor University}
  \city{Bremen}
  \country{Germany}
}

\begin{abstract}
 Interactive Theorem Proving was repeatedly shown to be fruitful when combined with Generative Artificial Intelligence. This paper assesses multiple approaches to Rocq generation and illuminates potential avenues for improvement. We identify retrieval-based premise selection as a central component of effective Rocq proof generation and propose a novel approach based on a self-attentive embedder model. The evaluation of the designed approach shows up to 28\% relative increase of the generator's performance. We tackle the problem of writing Rocq proofs using a multi-stage agentic system, tailored for formal verification, and demonstrate its high effectiveness. We conduct an ablation study and demonstrate that incorporating multi-agent debate during the planning stage increases the proof success rate by 20\% overall and nearly doubles it for complex theorems, while the reflection mechanism further enhances stability and consistency.
\end{abstract}

\keywords{Rocq, Coq, theorem proving, proof assistant, premise selection, agentic system, multi-agentic debate}

\newcommand{\BibTeX}{\rm B\kern-.05em{\sc i\kern-.025em b}\kern-.08em\TeX}

\makeatletter

\newcommand{\rocqstar}{%
  {\rmfamily R\kern-.05em o\kern-.05em c\kern-.05emq}%
  \textsuperscript{$\star$}%
}

\definecolor{darkgreen}{RGB}{18, 94, 34}
\newcommand{\thrtoken}[1]{\textcolor{darkgreen}{\texttt{#1}}}

\newcommand{\defhighlighter}[3][]{%
  \tikzset{every highlighter/.style={color=#2, fill opacity=#3, #1}}%
}

\defhighlighter{yellow}{.5}

\newcommand{\highlight@DoHighlight}{
  \fill [ decoration = {random steps, amplitude=1pt, segment length=15pt}
        , outer sep = -15pt, inner sep = 0pt, decorate
        , every highlighter, this highlighter ]
        ($(begin highlight)+(0,8pt)$) rectangle ($(end highlight)+(0,-3pt)$) ;
}

\newcommand{\highlight@BeginHighlight}{
  \coordinate (begin highlight) at (0,0) ;
}

\newcommand{\highlight@EndHighlight}{
  \coordinate (end highlight) at (0,0) ;
}

\newdimen\highlight@previous
\newdimen\highlight@current

\DeclareRobustCommand*\highlight[1][]{%
  \tikzset{this highlighter/.style={#1}}%
  \SOUL@setup
  \def\SOUL@preamble{%
    \begin{tikzpicture}[overlay, remember picture]
      \highlight@BeginHighlight
      \highlight@EndHighlight
    \end{tikzpicture}%
  }%
  \def\SOUL@postamble{%
    \begin{tikzpicture}[overlay, remember picture]
      \highlight@EndHighlight
      \highlight@DoHighlight
    \end{tikzpicture}%
  }%
  \def\SOUL@everyhyphen{%
    \discretionary{%
      \SOUL@setkern\SOUL@hyphkern
      \SOUL@sethyphenchar
      \tikz[overlay, remember picture] \highlight@EndHighlight ;%
    }{%
    }{%
      \SOUL@setkern\SOUL@charkern
    }%
  }%
  \def\SOUL@everyexhyphen##1{%
    \SOUL@setkern\SOUL@hyphkern
    \hbox{##1}%
    \discretionary{%
      \tikz[overlay, remember picture] \highlight@EndHighlight ;%
    }{%
    }{%
      \SOUL@setkern\SOUL@charkern
    }%
  }%
  \def\SOUL@everysyllable{%
    \begin{tikzpicture}[overlay, remember picture]
      \path let \p0 = (begin highlight), \p1 = (0,0) in \pgfextra
        \global\highlight@previous=\y0
        \global\highlight@current =\y1
      \endpgfextra (0,0) ;
      \ifdim\highlight@current < \highlight@previous
        \highlight@DoHighlight
        \highlight@BeginHighlight
      \fi
    \end{tikzpicture}%
    \the\SOUL@syllable
    \tikz[overlay, remember picture] \highlight@EndHighlight ;%
  }%
  \SOUL@
}
\makeatother

\definecolor{mDarkTeal}{HTML}{23373b}
\definecolor{filegray}{RGB}{245,245,245}
\definecolor{stateclr}{RGB}{208,0,0} 

\newcommand{\ie}{\emph{i.e., }}

\crefname{part}{\S}{\S\S}
\crefname{chapter}{\S}{\S\S}
\crefname{section}{\S}{\S\S}
\crefname{subsection}{\S}{\S\S}
\crefname{figure}{fig.}{Fig.}

\definecolor{toolheader}{HTML}{DCEDF0}
\definecolor{toolborder}{HTML}{BDD7DA}

\tikzset{
  toolbox/.style={
    shape=rectangle split,
    rectangle split parts=2,
    rectangle split part fill={toolheader,white},
    draw=toolborder,
    rounded corners=4pt,
    inner ysep=1pt,
    inner xsep=6pt,
    text width=2.8cm,
    rectangle split part align={center,top},
    font=\sffamily\small
  }
}

\definecolor{llmbg}{HTML}{ddecee}
\definecolor{llmstroke}{HTML}{9ca3e3}

\tikzset{
  llm block/.style={
    draw=llmstroke,
    line width=1pt,
    fill=llmbg,
    rounded corners=10pt,
    inner xsep=0pt,
    inner ysep=5pt,
    align=center,
    font=\sffamily\bfseries
  }
}

\lstdefinestyle{semi}{
  language=coq,
  basicstyle=\ttfamily\linespread{1.15}\tiny,
  backgroundcolor=\color{gray!05},
  xleftmargin=0pt,
  xrightmargin=0pt,
  columns=fullflexible,
  frame=tlbr, 
  framerule=0pt,
}

\newcommand{\LLMNode}[3][]{%
  \node[llm block,#1] (#2) {
    \begin{tabular}{@{\hspace{6pt}}c@{\hspace{4pt}}}
      \includegraphics[width=1.8em]{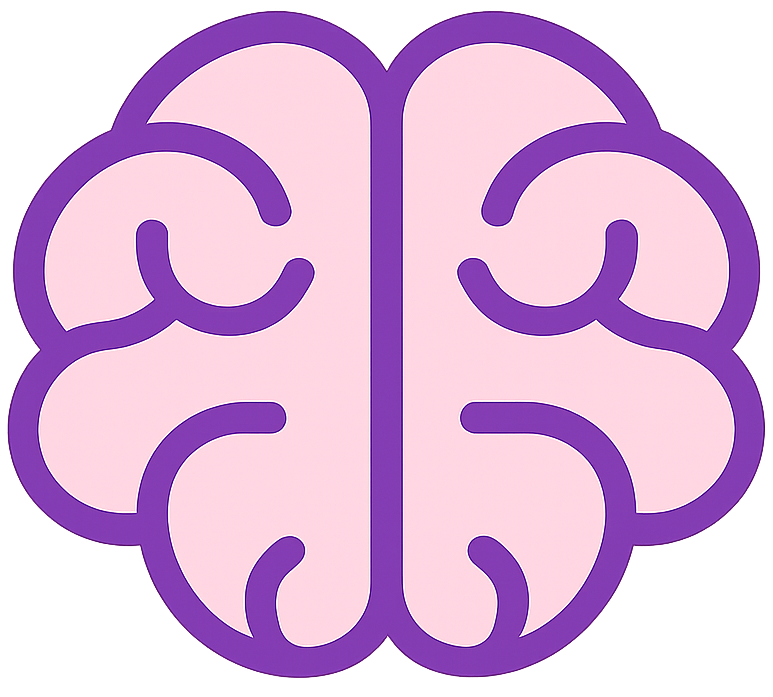}\\[4pt]
      #3
    \end{tabular}
  };
}

\newcommand{\LLMNodeSmall}[3][]{%
  \node[llm block,#1] (#2) {
    \begin{tabular}{@{\hspace{6pt}}c@{\hspace{4pt}}}
      \includegraphics[width=1em]{img/llm-logo-pink.png}\\[1pt]
     \sffamily\tiny #3
    \end{tabular}
  };
}

\newcommand{\marknumtext}[2][]{%
  \tikz[baseline=(C.base)]{
    \node[
      circle,
      draw=numcolor,
      fill=numcolor,
      inner sep=1pt,
      minimum size=1em,
      text=white,
      font=\small\sffamily\bfseries,
      #1
    ] (C) {#2};
  }%
}

\definecolor{numcolor}{HTML}{0D3B66}
\newcommand{\marknum}[3][]{%
  \node[
    circle,
    draw=numcolor,
    fill=numcolor,
    inner sep=1pt,
    minimum size=1em,
    text=white,
    font=\tiny\sffamily\bfseries,
    #1
  ] at #2 {#3};
}

\definecolor{lightgray}{rgb}{.9,.9,.9}
\definecolor{darkgray}{rgb}{.4,.4,.4}
\definecolor{purple}{rgb}{0.65, 0.12, 0.82}

\lstdefinestyle{coqstyle}{
  language=coq,
  basicstyle=\ttfamily\scriptsize,
  backgroundcolor=\color{filegray},
  frame=single,
  showstringspaces=false,
  xleftmargin=5.5pt,
  xrightmargin=5.5pt,
  framexleftmargin=2pt,
  framexrightmargin=2pt,
}

\lstdefinelanguage{javascript}{
  keywords={abstract, any, as, boolean, break, case, catch, class, console, const, continue, debugger, declare, default, delete, do, else, enum, export, extends, false, finally, for, from, function, get, if, implements, import, in, infer, instanceof, interface, keyof, let, module, namespace, never, new, null, number, object, package, private, protected, public, readonly, require, return, set, static, string, super, switch, symbol, this, throw, true, try, type, typeof, undefined, unique, unknown, var, void, while, with, yield},
  keywordstyle=\color{blue}\bfseries,
  ndkeywords={class, export, boolean, throw, implements, import, this},
  ndkeywordstyle=\color{darkgray}\bfseries,
  identifierstyle=\color{black},
  sensitive=false,
  comment=[l]{//},
  morecomment=[s]{/*}{*/},
  commentstyle=\color{purple}\ttfamily,
  stringstyle=\color{red}\ttfamily,
  morestring=[b]',
  morestring=[b]"
}

\lstdefinestyle{jsstyle}{
  language=javascript,
  basicstyle=\ttfamily\scriptsize,
  extendedchars=true,
  backgroundcolor=\color{filegray},
  frame=single,
  showstringspaces=false,
  xleftmargin=5.5pt,
  xrightmargin=5.5pt,
  framexleftmargin=2pt,
  framexrightmargin=2pt,
  captionpos=b
}

\newenvironment{acmstylefigure}
  {%
    \begingroup
    \renewcommand{\sfdefault}{phv}%
    \linespread{1.0}%
  }
  {%
    \endgroup
  }

\newcolumntype{C}{>{\centering\arraybackslash}m{1.4cm}}

\newcommand{\val}[2]{%
  \makebox[0pt][c]{#1\,{\textcolor{gray}{$\pm$\,#2}}}%
}

\begin{document}

\pagestyle{fancy}
\fancyhead{}

\maketitle 

%%%%%%%%%%%%%%%%%%%%%%%%%%%%%%%%%%%%%%%%%%%%%%%%%%%%%%%%%%%%%%%%%%%%%%%%

\section{Introduction}

In recent years, the advent of Generative Artificial Intelligence (AI) has accelerated the process of developing new software. However, there are studies~\cite{perry2022users} showing that users who use AI assistants tend to introduce more bugs and vulnerabilities into their code, compared to those who write code on their own. Formal software verification could help mitigate the issue of bugs and security flaws, as it ensures that the software operates correctly and reliably in compliance with the given specification. Under the assumption of a well-formed specification, formal verification provides strong guaranties and an acceptance criterion for the generated code. Interactive Theorem Prover (ITP) is a software tool that assists the user with the development of formal specifications and proofs. To date, there exist several ITPs, such as Rocq (formerly Coq)~\cite{coq}, Lean~\cite{lean}, Agda~\cite{agda}, Isabelle~\cite{isabelle}, and others. Rocq is a mature ITP, which has experienced more than 30 years of continuous development and improvement. Rocq has an extensive track record of high-impact projects. For example, Rocq was used to verify the correctness of the CompCert C compiler~\cite{compcert}, the only C compiler, in which an extensive study found no bugs~\cite{yang2011finding}.

Verifying software has always been a rigorous process requiring significant time and human effort. To mitigate this, a number of approaches have been developed to automate theorem proving in Rocq. These approaches build upon the underlying interactive proof paradigm of Rocq, where proofs are constructed incrementally using so-called \emph{tactics}. Tactics serve as elementary building blocks that allow the user to manipulate the \emph{proof state} --- a data structure containing the current goal and its context. Each applied tactic transforms the proof state, reducing the original task into simpler subgoals that can be solved recursively. Most solutions implement tactic-prediction approaches and employ beam search or a similar algorithm to navigate the search space. Tactician~\cite{tactician} is a kNN-based approach, which does similarity-based retrieval of tactics used in similar states. CoqGym~\cite{coqgym} and Proverbot9001~\cite{proverbot9001} use Recurrent Neural Networks (RNNs), Graph2Tac~\cite{rute2024graph2tac} proposed a novel graph-based neural tactic prediction. \citet{copra} and \citet{coqpilot} instead build generation pipelines around general-purpose, cloud-hosted LLMs, so that no heavy computations occur on the user's machine. CoqPilot~\cite{coqpilot}, along with that, contributes a benchmarking framework and allows seamless integration of standalone tools into the workflow of Rocq's user.

Many approaches call attention to the importance of premise selection, \ie retrieving useful context information to advance generation. \citet{yang2023leandojo} introduced LeanDojo, a retrieval-augmented prover in Lean that significantly improves over non-retrieval baselines. \citet{thompson2024rango} present the Rango tool and report state-of-the-art performance on the CoqStoq benchmark, automatically synthesizing complete proofs for 32\% of the theorems. The work highlights how strongly the well-formed context contributes to the success of Rango. Moreover, they show that \emph{proof retrieval} is the most performant mechanism for premise selection. The proof retriever selects relevant previously completed proofs from the current project and provides them as references to the model. According to the evaluation, Rango proved 47\% more theorems than the variant without a proof retriever. However, their retrieval mechanism assumes that two textually similar statements have proofs relevant to each other. In this work, we demonstrate that this assumption oversimplifies the relationship between statements and proofs and introduce a novel embedding model for Rocq statements. It is trained to predict the similarity between their proofs and achieves up to a 28\% relative improvement on the evaluation set.

Another promising direction in generative theorem proving that we have identified is Agentic Systems. Research by \citet{coqpilot} shows that current Rocq generation methods mostly struggle with complex reasoning tasks. Approaches that perform proof search on top of a tactic generator slow down dramatically as theorem complexity grows, since searching for longer proofs becomes exponentially harder due to the explosion of the underlying search space~\cite{wu2025advancingautomatedtheorem}. Other neural methods, which apply LLMs, suffer from the same problem due to the inability of the model to handle complex reasoning tasks~\cite{jiang2024peek}. Agentic systems are known to address these problems; however, to our knowledge, there were close to no attempts to build an autonomous agentic system for an ITP\@. We build an extensive Model Context Protocol (MCP) server for Rocq and implement an autonomous Agentic System over it, utilizing various problem-specific solutions, such as multi-agent debate. We conduct an evaluation and show that our agentic system strongly outperforms all other previously benchmarked solutions in the CoqPilot's work, raising the ratio of successfully proven theorems from 51\% to 60\%.

\subsection{Contributions}\label{sec:contributions}
The main contributions of this paper are the following.

\textbf{RocqStar proof retriever}\; We propose a novel approach for premise selection in Rocq. Rocq suffers from the data-scarcity problem that is common to most ITPs. Aggregating the largest publicly available repositories, one could expect to collect roughly 300 million tokens of Rocq, and about the same for Lean.
In contrast, open-source Python corpora easily exceed 100 billion tokens. To tackle this issue we contribute a convenient standalone tool \emph{BigRocq} to extract additional data from Rocq code, utilizing the nature of Rocq's system and the intermediate states of the proof.
BigRocq bridges the gap between Automated Generation and Rocq's ecosystem. Using BigRocq, we mine a dataset of 76,524 statements with corresponding proofs from 4 big projects and train a self-attentive embedder model, which learns to predict how close the proofs of given statements will be.
In addition, we provide a pipeline to reproduce such embeddings for an arbitrary project, which offers even better results.
We integrate the solution as a new retrieval approach for selecting context theorems in CoqPilot and evaluate it using CoqPilot's benchmarking infrastructure. Compared to the baseline text similarity-based ranker, we show an improvement of 28\% on the evaluation set.
The BigRocq tool, the training dataset, and the code for training the embedder model are available at \url{https://github.com/JetBrains-Research/rocqstar-rag}.
The embedder model's checkpoint is available at \url{https://huggingface.co/JetBrains-Research/rocq-language-theorem-embeddings}.

\textbf{RocqStar agentic system}\; Addressing the lack of research on applying agentic systems to ITPs, we build an autonomous system for generating Rocq proofs. A custom MCP server built on top of \texttt{coq-lsp}~\cite{emiliocoqlsp} handles the interaction with Rocq; its source code is available at \url{https://github.com/JetBrains-Research/rocqstar-agentic-system}. Our approach follows a structured process consisting of \emph{planning}, \emph{execution}, and \emph{reflection}. An ablation study shows that while naive planning has limited impact, effective planning based on the Multi-Agent Debate (MAD) framework plays a crucial role. Specifically, it yields a 20\% relative improvement in the overall proof success rate and nearly doubles the success rate on complex theorems with longer reference proofs (33\% vs.~17\%). Additionally, we demonstrate the benefits of the reflection mechanism, which improves the overall proof success rate from 48\% to 66\% and more than quadruples success on complex theorems; see \Cref{sec:ablationstudy} for details. The evaluation results show that the RocqStar system solves up to 60\% of the theorems from the CoqPilot dataset. It is implemented using Koog\footnote{Koog\: \url{https://docs.koog.ai}}, an open-source JetBrains framework that offers a type-safe Kotlin DSL for building AI agents with structured workflows and tool interaction. The source code of the agent is available at \url{https://github.com/JetBrains-Research/rocqstar-agentic-system}.

The remainder of the paper is organized as follows.
\Cref{sec:retrieval} describes our Similarity-Driven Retrieval mechanism.
\Cref{sec:agent} introduces the agentic system.
\Cref{sec:evaluation} presents an evaluation of
the retrieval component (\Cref{sec:evalranker}), the agent (\Cref{sec:evalagent}),
and an ablation study of the agentic system (\Cref{sec:ablationstudy}).
We describe the related work in \Cref{sec:relatedwork} and conclude in \Cref{sec:conclusion}.

\begin{figure*}[!t]
    \centering
    \begin{tikzpicture}[>=Stealth, node distance=2cm]
  \node[text width=2.5cm,
  fill=filegray,
        inner sep=8pt,
        inner ysep=-2pt,
        font=\ttfamily\scriptsize,
        anchor=west] (code) {
\begin{lstlisting}[language=coq, basicstyle=\scriptsize\ttfamily]
Theorem test : 
    forall n : nat, 
      n = 0 \/ n <> 0.
Proof.
    intros n.
    destruct n.
    - left; auto.
    - right; auto.
Qed.
\end{lstlisting}};

\node[right=0.7cm of code.east, anchor=west, inner sep=0pt, fill=filegray, every arrow/.style={-Stealth[scale=1], font=\scriptsize}] (tree) {
    \begin{tikzcd}[sep=small, every label/.append style = {font = \scriptsize}]
        &  &                                                            &  & \color{stateclr} s_2 \arrow[rr, "\texttt{left; auto.}"]  &  & \texttt{\color{stateclr} []} \\
        \color{stateclr} s_0 \arrow[rr, "\texttt{intros n.}"] &  & \color{stateclr} s_1 \arrow[rru, "\texttt{destruct n.}"] \arrow[rrd, "\texttt{destruct n.}" swap] &  &                                &  &     \\
        &  &                                                            &  & \color{stateclr} s_3 \arrow[rr, "\texttt{right; auto.}"] &  & \texttt{\color{stateclr} []}
    \end{tikzcd}
  };

  \node[right=0.7cm of tree.east, anchor=west] (substates) {
      \begin{tikzpicture}[node distance=0.5cm, >={Stealth[round]}]
        \node[fill=gray!10, inner sep=4pt, inner ysep=-4pt, text width=4.2cm, font=\ttfamily\scriptsize, anchor=west] (sub1) at (-0.5,0) {\begin{lstlisting}[language=coq, basicstyle=\scriptsize\ttfamily]
  Lemma s1 (n : nat) : n = 0 \/ n <> 0.
  Proof.
      destruct n.
      - left; auto.
      - right; auto.
  Qed.
          \end{lstlisting}};
        \node[fill=gray!10, inner sep=4pt, inner ysep=-4pt, text width=4.2cm, font=\ttfamily\tiny, anchor=west] (sub2) at (-0.5,-1.32) {
          \begin{lstlisting}[language=coq, basicstyle=\scriptsize\ttfamily]
  Lemma s2 : 0 = 0 \/ 0 <> 0.
  Proof. left; auto. Qed.
          \end{lstlisting}};
        \node[fill=gray!10, inner sep=4pt, inner ysep=-4pt, text width=4.2cm, font=\ttfamily\scriptsize, anchor=west] at (-0.5,-2.09) {
          \begin{lstlisting}[language=coq, basicstyle=\scriptsize\ttfamily]
  Lemma s3 (n : nat) : S n = 0 \/ S n <> 0.
  Proof. right; auto. Qed.
          \end{lstlisting}};
      \end{tikzpicture}
  };

  \draw[-Stealth, draw=mDarkTeal] (code.east) -- (tree.west);
  \draw[dashed, ->, draw=gray!60, opacity=0.6] (5.8, 0.0) -- (8.9, 0.7);
  \draw[dashed, ->, draw=gray!60, opacity=0.6] (6.8, 0.52) -- (8.9, -0.55);
  \draw[dashed, ->, draw=gray!60, opacity=0.6, bend right=20] (6.8, -0.98) to (8.9, -1.32);
\end{tikzpicture}
\caption{Processing theorems into trees; $s_i$ denotes a state}\label{fig:treefying}
\end{figure*}

\section{Similarity-driven Retrieval}\label{sec:retrieval}

A known problem in Retrieval Augmented Generation (RAG), applied to the domain of Interactive Theorem Proving (ITP), is \emph{premise selection}~\cite{urban2004mptp, irving2016deepmath}. Premise selection is the task of retrieving relevant facts from a given knowledge base to provide the model with the necessary context to advance the proof. However, \citet{Huang2025BreakingFC} and \citet{xu2024rereadingimprovesreasoninglarge} highlight that this context must be carefully curated, as the inclusion of irrelevant information degrades model performance. 

We distinguish two ways of doing premise selection in Rocq. \emph{Hint selection} --- given a context $C$ and a tactic with an unknown positional argument, e.g. \texttt{apply \_}, the task is to yield potential candidates for the argument. \emph{Proof selection}, in turn, given theorem statement $S$, focuses on choosing other statements with their respective proofs, so that their presence in the context of the generation request would help the model with the generation of the proof for statement $S$. Since the approach of applying general purpose models to proof generation is relatively new, most of the works~\cite{knntactician, neuralagda, thompson2024rango, yang2023leandojo} on premise selection in Rocq and other ITPs focused on hint selection. However, \citet{thompson2024rango} and \citet{coqpilot} show that even a baseline proof selection significantly boosts the model's capabilities and is stronger than hint selection. The baseline proof selection presented in both works \cite{thompson2024rango, coqpilot}, given the target statement $s_*$ and a database of already proven theorems $\left[s_0, p_0\right], \dots, \left[s_n, p_n\right]$ (where $s_i$ denotes a theorem statement and $p_i$ its respective proof), chooses theorems, statements of which have the maximum similarity to the target one. Similarity is defined by the BM-25 information retrieval technique~\cite{bm25} or the Jaccard similarity index. That results in packing the generator's context with theorems, based on how similar their statements are syntactically. 

We propose a retrieval mechanism that improves the performance of the generator compared to the described baseline. During the generation of the target proof for the statement $S$ we generally assume that the model benefits more from seeing similar proofs to the one that it needs to generate, rather than from seeing similar statements with proofs dissimilar from the target one. Evaluation of our retriever in \Cref{sec:evalranker} supports this supposition. One might assume that if statements $s_*$ and $s_i$ are similar, their respective proofs $p_*$ and $p_i$ are similar as well:
\begin{equation}
    \operatorname{similarity}(s_*, s_i) \;\Longrightarrow\; \operatorname{similarity}(p_*, p_i)
    \label{eq:stmt-proof-similarity}
\end{equation}
However, we show that this implication often \textbf{does not} hold. The heuristic of retrieving similar statements produces decent baseline results, but fails in complex cases, leaving room for improvement. We design our retrieval method to guide context selection based on the similarity of the proofs and show its practicality. 

Let us define the proof similarity $D_L$ as the Levenshtein edit distance computed over lists of tactics. Insertions and deletions of tactics have a unit cost, as in the standard Levenshtein formulation. The substitution cost between two tactics is proportional to the Levenshtein distance between their string representations. The resulting distance is normalized by the maximum proof length. 
\[
    p_i = \left[tac_{i_0}, \dots, tac_{i_m}\right], \quad l_i = |s_i|, \quad D_L(p_i,p_j) = \frac{\mathrm{Lev}(p_i,p_j)}{\max(l_i,l_j)}
\]
We conduct the following experiment to examine whether the relation in \Cref{eq:stmt-proof-similarity} holds in practice. Considering 1,855,701 pairs of theorems from the IMM project\footnote{IMM\: \url{https://github.com/weakmemory/imm}}, corresponding to all unordered pairs among 1,927 theorems, we compute correlations between statement similarities and respective proof similarities. In summary, BM25-based statement similarity shows a weak negative relationship with the Levenshtein-based proof distance (Pearson $r$ = –0.154, Spearman $\rho$ = –0.171). The code to reproduce these experiments could be found in the \emph{RocqStar-retriever} repository\footnote{RocqStar retriever: \url{https://github.com/JetBrains-Research/rocqstar-rag/tree/main/experiments}}.

To assess the issue of ineffective proof selection, we try to find a function $f(s_i, s_j)$ that correlates with the defined proof distance stronger than statement similarity does. In this work, we introduce a neural method that learns vector embeddings for Rocq theorem statements, training them so that the distance between any two vectors mirrors the similarity between the respective theorem's proofs.

\subsection{Dataset mining}
Along with other ITPs, Rocq struggles with data scarcity. To address this issue, we mine additional data from the Rocq code.
We utilize Rocq system's functionality, preprocess theorems, and transform sequential proof structures into trees.
\Cref{fig:treefying} illustrates this transformation process.
Since every node in such a tree is a valid state, we can automatically construct a proof for it by recursively following its subtree edges. Extracting all intermediate statements together with their proofs allows us to expand any given Rocq theorem dataset by a factor roughly proportional to the average proof length. In practice, the observed expansion is also affected by limitations of our extraction procedure. In our case, this resulted in an approximately fourfold increase in dataset size. The dataset format and further details are provided in Appendix~\ref{appendix:encoderdataset}.

We call the proposed tool \emph{BigRocq} and make it publicly available as a standalone component of our system. The idea of mining additional training data from the intermediate states of the ITP is not new; \citet{neuralagda} conducted analogous research for the Agda~\cite{agda} language. Similar research for Rocq also takes place~\cite{coqgym, rute2024graph2tac}; however, some of those works are highly dependent on the deprecated ways of communication with Rocq's compiler~\cite{coqgym} and do not support up-to-date versions of Rocq. In contrast, others implement similar ideas as a part of the training pipeline and do not allow for seamless reuse. Using BigRocq, we mine a total of 76,524 statements, collected from 344 files from 4 big Rocq projects. These projects are CompCert, IMM, Promising2Imm, and XMM. CompCert contributes large and heterogeneous proofs, while the remaining projects focus on weak-memory reasoning, which exhibits diverse proof styles across multiple abstraction levels. To avoid data leakage, we exclude theorems belonging to the evaluation datasets from training and only mine subgoals when augmenting a project.

\subsection{Modeling}\label{sec:emb-modeling}
In our work, we formulate the problem as a self-supervised contrastive representation learning task and train a self-attentive embedder model~\cite{lin2017structured}. Given a dataset $\mathcal{T} = {(s_i, p_i)}$, where $s_i$ is a Rocq statement, $p_i$ is its corresponding proof, and $\mathcal{S}$ denotes the set of all statements appearing in $\mathcal{T}$, together with a similarity function $f(\operatorname{proof}_i, \operatorname{proof}_j)$ defined between two proofs, we aim to learn a function
\[
r : \mathcal{S} \times \mathcal{S} \to \mathbb{R},
\]
which takes two statements $s_i, s_j \in \mathcal{S}$ as inputs and outputs a score approximating the similarity of their respective proofs $p_i, p_j$. In other words, $r(s_i, s_j) \approx f(p_i, p_j)$. This formulation allows the ranker $r$ to assign scores to candidate statements relative to a target statement, thereby guiding retrieval towards those whose proofs are most likely to be useful.

In \Cref{sec:evaluation}, we evaluate the performance of the proposed model in the following task. Given a target statement $s_*$ and a set of proven theorems $\mathcal{T}$, we aim to select $k$ premises from $\mathcal{T}$ to be used as context for generating a proof of $s_*$. We take the $k$ most relevant theorems, according to the ranker $r$.
\[
\operatorname{Top}_k(r, s_*) = \underset{(s_i, p_i)\in\mathcal{T}}{\arg\operatorname{top}_k}\, r(s_i, s_*)
\]
We say that statement $s_*$ is solved with the use of the ranker $r$ if the generator $g$ produces a valid proof for $s_*$ given the premises selected by $r$.
\[
\operatorname{Solve}(s_*, r, g) = 
\begin{cases}
1, & g\bigl(\operatorname{Top}_k(r, s_*),\, s_*\bigr) \;\text{is a valid proof}, \\
0, & \text{otherwise}.
\end{cases}
\]
Finally, the quality of the ranker is estimated by the number of theorems in the evaluation set that can be solved using $r$ in combination with a given generator $g$.

One of the difficulties we encountered during training is a U-shaped distribution of proof distances over randomly sampled theorem pairs. Specifically, when plotting the frequency of theorem pairs against their proof distance, most pairs cluster either at very small distances, corresponding to short, highly similar proofs, or at very large distances, corresponding to largely unrelated proofs. As a result, pairs with intermediate distances are underrepresented, creating a gap that hinders effective training. To mitigate this, we define a modified proof distance that combines the previously introduced in \Cref{sec:retrieval} distance measures with an additional similarity term and injected noise for robustness:
\begin{align*}
    \mathrm{proof\_distance}(p_i,p_j)
    &= \alpha\,D_L(p_i,p_j)+(1-\alpha)\,D_J(p_i,p_j) + \gamma \\ 
    D_J(p_i,p_j) &= 1 - \frac{|p_i \cap p_j|}{|p_i \cup p_j|}
\end{align*}
The coefficient $\alpha = 0.7$ was chosen heuristically based on the distribution plot and yielded the best performance in experiments. The noise $\gamma$ is taken from $\mathcal{U}(-1e-3, +1e-3)$. 

As we have already explained in \Cref{sec:retrieval}, statement similarity is a poor choice of $r$, as it shows a low correlation with the target function $\mathrm{proof\_distance}$. 
However, it still provides a strong baseline: in practice, similar theorems occasionally have similar proofs. Accordingly, our approach builds upon statement encoders, fine-tuning them to better align with the underlying proof structure. We fine-tune Microsoft's 108-million-parameter encoder CodeBert~\cite{feng2020codebert}, originally pretrained on a combined corpus of programming and natural language texts. We also experimented with \texttt{gte-modernbert-base}\footnote{gte-modernbert-base: \url{https://huggingface.co/Alibaba-NLP/gte-modernbert-base}} as the base model, but it did not yield notable improvements. As shown in \Cref{sec:evalranker}, the encoder without post-training performs on par with the Jaccard-similarity baseline, indicating that the unadapted model relies primarily on surface-level syntactic similarity rather than proof-related semantics. Our goal, therefore, is to adapt the encoder to capture this deeper semantic relation.

To achieve this, we train the model using the InfoNCE~\cite{infonce} loss. In our setting, the distribution of proof distances is imbalanced even after normalization. InfoNCE naturally handles this case by contrasting a limited number of positives against a set of negatives, ensuring that informative gradients are maintained. In particular, given a statement $s$, during dataset preprocessing we compute distances to other samples. 
We then mark a pair as positive if the distance between their proofs is less than a threshold $\tau_{pos}$, and mark it as negative if it is greater than $\tau_{neg}$. 
Given the hyperparameter $k_{\operatorname{neg}}$ and sets of positive and negative pairs $P^+_s$ and $P^-_s$, we compute a per-statement loss term $\mathcal{L}_s$ as follows:
\begin{gather*}
\mathcal{L}_{s}
 =
-\log
\frac
{\exp \bigl(\cos(z_s,z_{p}) / T\bigr)}
{\exp \bigl(\cos(z_s,z_{p}) / T\bigr) +
       \displaystyle\sum_{j=1}^{k_{\operatorname{neg}}}
       \exp \bigl(\cos(z_s,z_{n_j}) / T\bigr)}
\end{gather*}
where $\cos$ is a cosine similarity between $\ell_2$-normalized embeddings of statements, and $p\in P^+_{s}$, $n_j\in P^-_{s}$.
On average, we observed smoother convergence for higher values of $k_{\operatorname{neg}}$, which is consistent with findings by \citet{chen2022we}.
However, due to hardware limitations, we selected $k_{\operatorname{neg}} = 100$ as a practical trade-off between convergence stability and computational cost.

Despite the adjustment of $\mathrm{proof\_distance}(\cdot)$, during training we experienced the problem of the model converging too quickly on \textit{``easy''} negatives---pairs, whose proofs (and typically their statements) are already far apart in the raw distance space. To keep informative gradients flowing, we add \emph{hard negative} pairs; with some probability we treat a pair of statements as negative if $\tau_{\operatorname{hardneg}} \leqslant \operatorname{sim}(\text{proof}_a, \text{proof}_b)\leqslant \tau_{\operatorname{neg}}$. Introduction of negative samples helped to stabilize the training process; we have observed a less steep training curve and better generalization overall. Other training hyperparameters are listed in Appendix~\ref{appendix:encodertraining}.

\begin{acmstylefigure}
\begin{figure*}[!t]
    \centering

    \begin{tikzpicture}[node distance=4mm and 3mm,>=Stealth]
        \node[toolbox] (state) {
            \texttt{Target theorem} \nodepart{second}
            \begin{lstlisting}[style=semi]
b : bool
negb (negb b) = b.
            \end{lstlisting}
            };
        
                \node[toolbox, below=of state] (code) {
            \texttt{Premises} \nodepart{second}
            \begin{lstlisting}[style=semi]
Theorem plusn0: n : nat,
    (n + 1) =? 0 = false.
Proof. ... Qed.
            \end{lstlisting}
            \begin{lstlisting}[style=semi]
Theorem add_comm: b c,
    andb b c = andb c b.
Proof. ... Qed.
            \end{lstlisting}
            \centering $\vdots$
            \begin{lstlisting}[style=semi]
Theorem plus0: n : nat,
    0 + n = n.
Proof. ... Qed.
            \end{lstlisting}
          };
        
          \begin{scope}[on background layer]
            \node[
              name=outerLeft,
              fit=(state) (code),
              draw=gray!60,
              dashed,
              rounded corners=8pt,
              inner sep=8pt
            ] {};
          \end{scope}

          \marknum[anchor=west,xshift=2pt]{([xshift=-5pt, yshift=-3pt]outerLeft.north west)}{1};
      
        \begin{scope}[shift={($(outerLeft.east)+(1.5cm,1.64cm)$)}]
            \tikzset{node distance=12mm and 30mm, >=Stealth}
            \LLMNode{pro}{Pro LLM}
            \LLMNode[right=of pro]{con}{Con LLM}
            \LLMNode[below=of $(pro)!0.5!(con)$]{judge}{Judge LLM}
          
          \draw[->, gray, shorten >=1pt, shorten <=1pt]
            ([yshift=10pt]pro.east) 
                -- node[midway, above=-2pt, font=\small] {\texttt{Initial plan}} 
            ([yshift=10pt]con.west);
            
            \coordinate (A1) at ([yshift=3pt]pro.east);
            \coordinate (B1) at ([yshift=3pt]con.west);
            \coordinate (A2) at ([yshift=-12pt]pro.east);
            \coordinate (B2) at ([yshift=-12pt]con.west);
        
            \coordinate (A21) at ([yshift=-25pt]pro.east);
            \coordinate (B21) at ([yshift=-25pt]con.west);
          
            \draw[<-,gray,shorten >=1pt,shorten <=1pt]
              (A1) -- node[below,font=\small] {\texttt{Args against}} (B1);
            \draw[->,gray,shorten >=1pt,shorten <=1pt]
              (A2) -- node[below,font=\small] {\texttt{Args pro}} (B2);

            \begin{scope}[]
              \node[
                name=debate,
                fit=(A1)(B1)(A21)(B21),
                draw=red!70,
                dashed,
                rounded corners=4pt,
                inner sep=4pt
              ] {};
            \end{scope}
        
            \draw[->, red, bend right=25]
                ([xshift=2mm]debate.south west) to (judge.west);
        
            \node[font=\sffamily\tiny\bfseries, fill=white, text=red!70] 
              (krounds) 
              at ($(A21)!0.5!(B21)-(0,1mm)$)
              {\texttt{m rounds}};
          
            \node[font=\sffamily\large\bfseries] 
              (title) 
              at ($(pro)!0.5!(con)+(0,12mm)$) 
              {Planning Stage};
        
            \begin{scope}[on background layer]
              \node[
                name=outer,
                fit=(title)(pro)(con)(judge),
                draw=gray!60,
                dashed,
                rounded corners=8pt,
                inner sep=8pt
              ] {};          
            \end{scope}

            \marknum[anchor=west,xshift=2pt]{([xshift=-5pt, yshift=-3pt]outer.north west)}{2};
        
            \begin{scope}[]
                \node[
                    name=westcircle,
                  circle,
                  minimum size=5pt,
                  inner sep=0pt,
                  line width=0.5pt,
                  fill=llmbg,
                  draw=llmstroke
                ] at ([yshift=15pt]outer.west) {};
                \node[
                  name=eastcircle,
                  circle,
                  minimum size=5pt,
                  inner sep=0pt,
                  line width=0.5pt,
                  fill=llmbg,
                  draw=llmstroke
                ] at (outer.east) {};
              \end{scope}

              \draw[->,gray, bend left=20]
              (state.east) to (westcircle.north);
        
              \draw[->, gray, bend right=10]
              (judge.east) to 
                node[midway, sloped, below, font=\tiny] {\texttt{Winning plan}}
            (eastcircle.west);
        \end{scope}

        \begin{scope}[shift={($(outer.east)+(1.3cm,1.46cm)$)}]
            \LLMNode{planscoring}{\sffamily\scriptsize Plan scoring \\ \sffamily\scriptsize LLM}
        \end{scope}

        \draw[->,gray, bend left=10]
              (eastcircle.north) to 
                node[midway, sloped, above=-1mm, font=\tiny] {\texttt{k plans}}
              (planscoring.west);

        \begin{scope}[shift={($(outer.east)+(1.31cm,-1.65cm)$)}]
            \LLMNode{critic}{\sffamily\footnotesize Critic LLM}
        \end{scope}

        \begin{scope}[shift={($(outerLeft.east)+(1.3cm,-2.5cm)$)}]
            \tikzset{node distance=10mm and 14mm, >=Stealth}

            \node (cossim) {
                \includegraphics[width=3em]{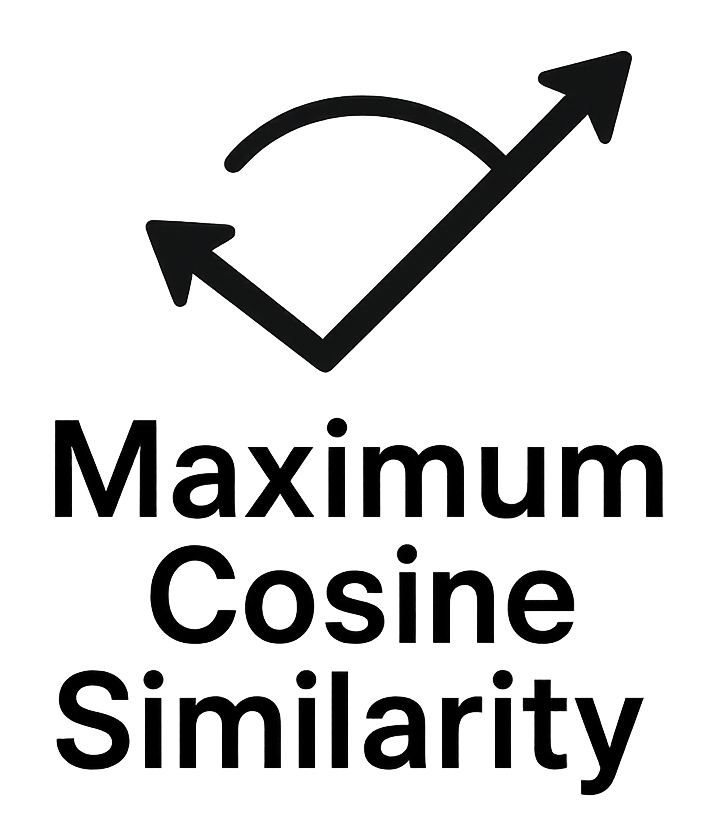}
            };

            \node[right=of cossim] (mcp) {
                \includegraphics[width=4em]{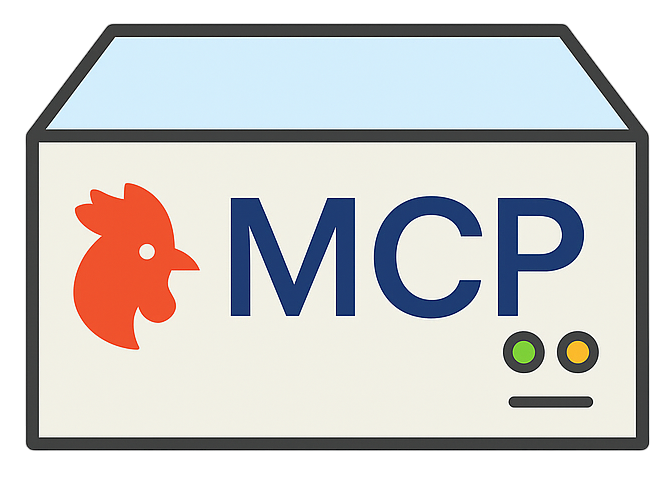}
            };
            
            \LLMNodeSmall[right=of mcp]{executor}{\sffamily\tiny Executor LLM}

            \draw[->,gray, bend left=20, shorten >=3pt]
              ([yshift=5pt]mcp.east) to ([yshift=5pt]executor.west);
              
            \draw[->,gray, bend left=20, shorten <=3pt]
              ([yshift=-5pt]executor.west) to ([yshift=-5pt]mcp.east);

            \draw[<->,gray, shorten >=3pt, bend right=5]
              ([yshift=-10pt]mcp.north east) to (critic.west);

            \draw[->,gray]
              (cossim.east) to 
              node[midway, sloped, above=-0.6mm, font=\tiny] {\texttt{analyzed}}
              (mcp.west);

            \draw[->,gray, bend left=20]
              (mcp.north west) to 
              node[midway, sloped, above=1mm, font=\tiny] {\texttt{Available tools}}
              (westcircle.south);

            \begin{scope}[on background layer]
                \node[
                  name=outer,
                  fit=(cossim)(mcp)(executor),
                  draw=gray!60,
                  dashed,
                  rounded corners=8pt,
                  inner ysep=4pt,
                  inner xsep=8pt,
                ] {};
                
                \node[
                    name=northcircle,
                    circle,
                    minimum size=5pt,
                    inner sep=0pt,
                    line width=0.5pt,
                    fill=llmbg,
                    draw=llmstroke
                  ] at ([xshift=20pt]outer.north) {};

                \marknum[anchor=west,xshift=2pt]{([xshift=-5pt, yshift=-3pt]outer.north west)}{3};
              \end{scope}

              \coordinate (P) at ([xshift=-10pt]planscoring.south);
                \coordinate (Q) at (northcircle.north east);

                \coordinate (LabelPos) at 
                    ($ (P) ! 1cm ! (Q) $);

                \draw[->,gray] (P) to (Q);

                \node at ([xshift=26pt, yshift=3pt]LabelPos)
                    [fill=filegray, font=\tiny, align=center] 
                    {\texttt{Plan 1: 0.95}\\\dots\\\texttt{Plan k: 0.36}};

            \draw[->,gray, shorten <=3pt, bend right=10]
            (critic.west) to 
            node[pos=0.4, sloped, above=-0.8mm, font=\tiny] {\texttt{Upd plan}}
            (northcircle.north east);

            \draw[->,gray, bend right=30]
            (outer.east) to 
            ([xshift=10pt]critic.south);

            \coordinate (rg) at (outer.east);
            \coordinate (lc) at (critic.south);
            \coordinate (labelpos2) at 
                    ($ (rg) ! 1cm ! (lc) $);

            \node at ([xshift=10pt, yshift=-15pt]labelpos2)
            [text=red!70, font=\tiny, fill=white, align=center] 
            {\texttt{if fails} \\ \texttt{often}};
            
            \draw[->,gray]
              ([yshift=-45.5pt]code.east) to 
              node[midway, sloped, above=-0.6mm, font=\tiny] {\texttt{Embedder}}
              ([yshift=0pt]cossim.west);

            \draw[->,gray]
              (state.south east) to 
              node[midway, sloped, above=-0.6mm, font=\tiny] {\texttt{Embedder}}
              ([xshift=-5pt]cossim.north);
              
        \end{scope}
      \end{tikzpicture}
    \caption{Agentic pipeline with RocqStar retriever}
    \label{fig:pipeline}
\end{figure*}
\end{acmstylefigure}

\section{Agentic System}\label{sec:agent}
Agent-based approaches are broadly used in code generation and repair tasks~\cite{sweagent, autocoderover, repairagent}. Despite a large number of autonomous and semi-autonomous coding agents, they are not widely used in formal proofs generation and are not tailored to the Rocq specifics. To address this, we have implemented a RocqStar agentic system.

To allow interaction between the agent and Rocq's system, we develop a REST API server that provides a set of tools that are useful during the execution. We apply our domain knowledge and construct these tools to bring an agent-driven proving process as close as possible to a human-driven one. Examples of allowed function calls include checking validity of proofs, retrieving the valid prefix of given proof, gathering additional information about available entities in the context, and interacting with the context via performing commands like \texttt{Print ?a} to identify the type of an argument or \texttt{Search ?exp} to search for defined terms by a pattern. Toolset is described in detail in Appendix~\ref{appendix:toolset}. Interaction with Rocq's system is carried out through its language server, \texttt{coq-lsp}~\cite{emiliocoqlsp}. To conform with a commonly used Model Context Protocol (MCP) and allow seamless agent interaction with the environment through tools, we implement an MCP server that wraps the REST API server. Among the provided tools, the most important is the proof-checking tool. It not only verifies whether a proof is valid but, in case of an error, returns detailed diagnostic information: the error message, its exact location, the valid prefix preceding the error, and the remaining goals after that prefix. This functionality allows the agent to maintain awareness of the current proof state and leverage partial proof progress.

\subsection{Agent Logic}

The input to the agent is presented as a target theorem without a proof and a file where it was declared,
see box $\marknumtext{1}$ of \Cref{fig:pipeline}.
Agent's pipeline is logically split into two main stages: \emph{planning} and \emph{execution}.
In the planning phase, multiple language models rigorously work out the strategy for the further implementation.
During execution agents follow the plan aiming to generate the correct proof.

\textbf{Planning Stage}\; We employ the idea of multi-agent debates~\cite{liang2023multiagenticdebate} to generate a strategy for proving the given theorem. Specifically, two LLMs engage in a discussion: one proposes and defends an initial plan (\emph{pro} LLM), while the other critiques it (\emph{con} LLM); see box~$\marknumtext{2}$ in \Cref{fig:pipeline}. After several debate rounds, the entire message history is passed to a \emph{judge} LLM, which determines the winner and produces the final plan. By repeating this procedure, we generate $k$ candidate strategies. These are then evaluated by a \emph{plan scoring} LLM that assigns each a numerical score (the higher, the better). Finally, the top-$l$ plans are selected and forwarded to the \emph{Execution Stage}; see box~$\marknumtext{3}$ of \Cref{fig:pipeline}.

\textbf{Execution Stage}\; For each of the selected plans, we run an \textit{executor} agent that follows it step by step, invoking tools from the provided tool set --- proof checker, context–inspection queries, search commands, and others, as atomic actions. Through these tool calls, the agent interacts with the environment via the MCP server.
In addition to this iterative execution, we employ a reflection mechanism that monitors the progress of the proof and adjusts the strategy when necessary. We track how many consecutive erroneous proof attempts occur, and once this number exceeds a predefined threshold (set to five during evaluation), a \textit{critic} model is called to assess the current proof state and identify deviations from the intended strategy.
After that, we retrieve theorems along with their proofs, whose top-level goals are similar to the currently remaining goal, according to the cosine similarity between their RocqStar-ranker embeddings. We prompt the LLM to explain which tactic sequences could be helpful to finish our proof. We gather the generated criticism and send it to the \textit{replanner} LLM to refine the current plan along with similar proofs and their analysis.
The replanner is a separate language model that revises the plan based on the critic's feedback and the retrieved examples. The whole message history is sent back to the \textit{executor} agent. During the execution of each plan, $n$ tool calls are allowed. If valid proof is not found after $n$ tool calls, we denote the plan as failed. In this case, we ask a \textit{plan failure summarizer} LLM to generate a short explanation of why the strategy execution failed and what happened during it. Then this summarized explanation is sent to the new execution stage with the next selected plan. This procedure is repeated until the correct proof is found or there are no more strategies to execute. For clarity, Figure~\Cref{fig:pipeline} presents a simplified view of the system and does not explicitly depict all auxiliary LLM components used in the pipeline.

\begin{table*}[!t]
  \centering
  \begin{tabular}{
    l
    *{3}{C}
    *{3}{C}
    *{3}{C}
  }
  \toprule
  \multicolumn{1}{l}{Group}
    & \multicolumn{3}{c}{$\mathbf{\leqslant 4}$}
    & \multicolumn{3}{c}{$\mathbf{5\,-\,8}$}
    & \multicolumn{3}{c}{$\mathbf{9\,-\,20}$} \\
  \cmidrule(lr){2-4} \cmidrule(lr){5-7} \cmidrule(lr){8-10}
  \multicolumn{1}{l}{Ranker}
    & Jaccard & ModernBert & \textbf{RocqStar}
    & Jaccard & ModernBert & \textbf{RocqStar}
    & Jaccard & ModernBert & \textbf{RocqStar} \\
  \midrule
  GPT-4o
    & \val{48\%}{5\%} & \val{44\%}{4\%} & \textbf{\val{51\%}{5\%}}
    & \val{18\%}{4\%} & \val{21\%}{5\%} & \textbf{\val{25\%}{3\%}}
    & \val{11\%}{4\%} & \val{8\%}{4\%} & \textbf{\val{14\%}{5\%}} \\
  Claude 3.5
    & \val{58\%}{5\%} & \val{57\%}{3\%} & \textbf{\val{61\%}{4\%}}
    & \val{28\%}{5\%} & \val{30\%}{3\%} & \textbf{\val{36\%}{5\%}}
    & \val{16\%}{5\%} & \val{16\%}{5\%} & \textbf{\val{21\%}{5\%}} \\
  \bottomrule
  \end{tabular}
  \vspace{0.2cm}
  \caption{Model performance under different ablations across all evaluation sets}
  \label{tab:eval_ranker}
\end{table*}

\begin{table*}[t]
  \centering
  \begin{minipage}{0.48\textwidth}
    \centering
    \begin{tabular}{>{\raggedright\arraybackslash}p{3cm}>{\raggedleft\arraybackslash}p{0.85cm}>{\raggedleft\arraybackslash}p{0.87cm}>{\raggedleft\arraybackslash}p{0.9cm}>{\raggedleft\arraybackslash}p{0.85cm}}
      \toprule
      Reference proof length & $\leqslant 4$ & 5--8 & 9--20 & Total \\
      Group size & 131 & 98 & 71 & 300 \\
      \midrule
      OpenAI GPT-4o & 50\% & 26\% & 15\% & 34\% \\
      OpenAI o1     & 66\% & 31\% & 8\% & 41\% \\
      Deepseek R1   & 58\% & 29\% & 11\% & 37\% \\
      Claude 3.5 Sonnet & 73\% & 41\% & 27\% & 51\% \\
      LleMMa 7B     & 24\% & 11\% & 1\% & 15\% \\
      \midrule
      Tactician (\texttt{synth}) & 45\% & 23\% & 10\% & 29\% \\
      Rango          & 38\% & 18\% & 8\% & 25\% \\
      \midrule
      \textbf{RocqStar Agent} & \textbf{76\%} & \textbf{56\%} & \textbf{38\%} & \textbf{60\%} \\
      \bottomrule
    \end{tabular}
    \caption{Different Rocq generation methods via CoqPilot}
    \label{tab:coqpilot_experiments}
  \end{minipage}%
  \hfill
  \begin{minipage}{0.48\textwidth}
    \centering
    \begin{tabular}{>{\raggedright\arraybackslash}p{3cm}>{\raggedleft\arraybackslash}p{0.85cm}>{\raggedleft\arraybackslash}p{0.87cm}>{\raggedleft\arraybackslash}p{0.9cm}>{\raggedleft\arraybackslash}p{0.85cm}}
      \toprule
      Reference proof length & $\leqslant 4$ & 5--8 & 9--20 & Total \\
      Group size & 22 & 16 & 12 & 50 \\
      \midrule
      Full Agent & \textbf{91\%} & \textbf{56\%} & \textbf{33\%} & \textbf{66\%} \\
      Agent w/o MAD & 86\% & 44\% & 17\% & 56\% \\
      Agent w/o Planning & 86\% & 50\% & 17\% & 58\% \\
      Agent w/o {\rocqstar} retr. & 86\% & 50\% & 33\% & 62\% \\
      Agent w/o Reflection & 73\% & 44\% & 8\% & 48\% \\
      Claude 3.5 Sonnet & 86\% & 37\% & 8\% & 52\% \\
      \bottomrule
    \end{tabular}
    \caption{Ablation study of Multi-Agent Debate}
    \label{tab:ablation}
  \end{minipage}
\end{table*}

\section{Evaluation}\label{sec:evaluation}
To evaluate our approach, partially and as a whole, we use the CoqPilot benchmarking framework. We required a dataset with a large number of human-written theorems and proofs. To compare our solution to existing ones, we decided to re-use the dataset by \citet{coqpilot}.
It is limited to 300 theorems from the IMM project~\cite{podkopaev2019bridging}, which was suitable for us in terms of computational and financial costs. The theorems are partitioned into three groups, corresponding to the difficulty level. The length (in tactics) of the human-written reference proof of the theorem estimates its difficulty. The sizes of each group are chosen with respect to the initial distribution of proof lengths in the project. Final group sizes and length ranges of each group could be found in Table~\ref{tab:coqpilot_experiments}. The dataset is further restricted to theorems whose human-written reference proofs contain no more than 20 tactics, following the original setup of the CoqPilot benchmark. From now on, we will refer to the described dataset as the \emph{IMM-300} dataset. For smaller ablation studies we additionally prepared \emph{IMM-50}, a 50-theorem subset of IMM, constructed with the same procedure. No theorems from the dataset were present in the training set of the RocqStar ranker embedding model. Moreover, the training set only contained \emph{partial theorem goals}, \ie intermediate proof goals arising during proof execution, rather than the original top-level theorem statements. The split of both datasets into groups, details, and limitations are described in Appendix~\ref{appendix:evaldataset}. Computational and financial resources used for experiments are described in Appendix~\ref{appendix:expresources}.

\subsection{Retrieval Mechanism}\label{sec:evalranker}
We integrate our retrieval mechanism as a ranker into CoqPilot and evaluate it on the IMM-300 dataset with different models under the hood. To assess its performance, we compare our approach against two baselines: (i) an untrained embedder model (we use \texttt{gte-modernbert-base}, with \texttt{codebert-base} yielding comparable results), and (ii) a lexical similarity baseline based on the Jaccard index. In the latter, given a target theorem statement $s_*$ and a set of proven theorems $\left[s_0, p_0\right], \dots, \left[s_n, p_n\right]$, it ranks the theorems in descending order of $J(s_*,s_i)$, where $J(s_*,s_i)$ is the Jaccard-similarity index.
The statement is split into tokens by whitespaces, commas, etc.
Jaccard-similarity index is semantically almost the same as the BM-25 metric and produces the same numerical results.
For each theorem in the dataset, we take theorems within the same file, sort them using the ranker (Jaccard, ModernBert, or RocqStar, respectively), take the $k$ most relevant ones ($k$ is equal to 7 in our experiments) and send a request to the model to generate the completion.
The chosen theorems are being sent as a few-shot prompt.
Generation for each theorem is requested 12 times.
If the Rocq's system accepts any of the proofs, the theorem is considered solved.
The target metric in our evaluation is the ratio of solved theorems.
The evaluation results are presented in Table~\ref{tab:eval_ranker}. The reported values denote mean success rates, and the $\pm$ intervals correspond to the standard deviation across three independent runs.
  
As can be seen from Table~\ref{tab:eval_ranker}, our RocqStar ranker consistently outperforms both the Jaccard baseline and the untrained ModernBert encoder, demonstrating reliable gains across all evaluation groups. Most of the performance increase could be seen in the second group; we interpret these results as follows. For short theorems in the first group, the assumption that similar statements imply similar proofs often holds; therefore, all rankers perform comparably. For complex theorems from the third group, it rarely happens that two theorems have significantly similar proofs, resulting in less advancement space for the model. 

\subsection{Agentic System}\label{sec:evalagent}

We evaluate our agentic system on the IMM-300 dataset with the goal of solving as many theorems as possible. We use the strongest available Rocq proof–generating models, as measured by pass@12 on the CoqPilot benchmark. In addition, following the common principle of using a critic stronger than the executor~\cite{variationinverification}, we employ a more capable model for the planning and evaluation stages. Specifically, for all parts of the planning stage, we use the Claude~3.5 Sonnet model and perform two rounds of debate between actors.
Four plans are generated, and two are chosen for further execution.
During execution, 20 tool calls are allowed from the MCP server.
Additionally, after five proof-checking calls, the critic model (Claude 3.7 Sonnet) is invoked and analyzes whether a deviation from the initial plan has occurred.
We use Claude 3.5 Sonnet for the execution and re-planning, and Google Gemini Flash 2.0 for other tasks, due to the necessity of a big context.
Results of the evaluation are shown in Table~\ref{tab:coqpilot_experiments}.

As shown in Table~\ref{tab:coqpilot_experiments}, our agentic system outperforms other benchmarked models inside the CoqPilot framework. The strongest model so far was Claude 3.5 Sonnet, which achieves 51\% accuracy on the dataset, given 12 retries for each theorem. RocqStar agent achieves 60\%, showing vigorous improvement. In terms of financial costs, we estimate a run of an agent on one theorem at 1.3 US dollars, compared to 0.25 US dollars for 12 requests to the pure Claude 3.5 Sonnet in CoqPilot. Along with five language models invocated through the CoqPilot framework, we have compared our solution to other Rocq generation approaches, such as Tactician and Rango. On our IMM-300 dataset both solutions showed a result comparable to CoqPilot with OpenAI GPT-4o as the generator model.

\subsection{Ablation study}\label{sec:ablationstudy}
We conduct an ablation study to analyze the contribution of individual components of the agentic pipeline to the overall success rate. In particular, we investigate the effects of removing (1) the \emph{Multi-Agent Debate (MAD)} layer responsible for iterative plan refinement, (2) the \emph{Planning} stage entirely, (3) the \emph{RocqStar retrieval} module, and (4) the \emph{Reflection} mechanism responsible for forced retrieval, criticism, and replanning. All experiments are performed on the IMM-50 dataset, with all other system components kept unchanged. The results are summarized in Table~\ref{tab:ablation}.

\textbf{Planning}\; Considering that software-verification tasks cannot be solved ad hoc, without explicit planning, we measure how removing the MAD layer and reverting to single-pass planning affects the proportion of successfully proved theorems. We run two versions of the agent: one generates plans via MAD, and the other produces a single plan in one LLM call without further refinement.
Additionally, we include a configuration with the \emph{Planning} stage entirely disabled, where the executor immediately attempts to construct a proof without any plan. This comparison shows that an agent without planning performs nearly identically, or slightly worse, than one guided by a poor single-pass plan, suggesting that a suboptimal plan does not provide advantage. In contrast, multi-step planning via MAD yields a consistent improvement across all groups, confirming the importance of structured plan refinement.
An example of how MAD repairs a previously unsuccessful plan is presented in Appendix~\ref{appendix:planningmad}.

\textbf{RocqStar retrieval}\; To further assess the contribution of retrieval to the agentic pipeline, we evaluate the agent with its retrieval component replaced by a Jaccard-based baseline. This substitution results in a small but consistent decrease in performance, indicating that the proposed retrieval mechanism provides a meaningful, though moderate, improvement in the agentic setting. While the agentic architecture mitigates some of the limitations of weaker premise selection, effective retrieval remains an important component that contributes to more stable and reliable proof generation.

\textbf{Reflection}\; Finally, we disable the reflection mechanism that triggers forced retrieval and replanning after failed attempts. Without reflection, the agent loses its ability to recover from early mistakes, leading to a noticeable degradation of performance, especially on longer and more complex proofs. The result shows that reflection complements planning by enabling recovery from failed reasoning trajectories.

\section{Related Work}\label{sec:relatedwork}
Many Rocq generation methods improve generation using Retrieval Augmentation. Most of those works solve the hint selection problem~\cite{knntactician, thompson2024rango}, described in \Cref{sec:retrieval}. Those approaches build proofs tactic by tactic, retrieving relevant lemmas or definitions to use in the next step. The problem of retrieving existing proofs that can help advance generation is barely explored in the literature. CoqPilot~\cite{coqpilot} and Rango~\cite{thompson2024rango} address this problem by augmenting the generator’s context with previously proven theorems whose statements are similar to the target theorem being proved. Our work proposes a novel premise selection method and demonstrates improvements over the baselines used in prior work~\cite{coqpilot, thompson2024rango}.

In our multi-agentic system, we distribute responsibility across multiple agents, each responsible for a simpler subtask within the overall proof generation process. This decomposition, which separates high-level reasoning, planning, and execution, is a common design pattern in agentic systems. For example, \citet{li2024codetree} propose a task force split into Thinker, Solver, Critic, and Debug agents, while \citet{liang2023multiagenticdebate} introduce a multi-agent debate framework that encourages divergent reasoning in complex tasks. We show that explicitly separating planning from execution is essential for the formal verification pipeline. Theorem-proving demands a clear and high-level picture of the proof before executing any code. Running a multi-agent debate at the planning stage ensures rigorous evaluation of different approaches before interacting with Rocq's system. We produce several plans for further execution. In a manner similar to \citet{islam2024mapcoder}, we assign scores to plans and run them in the order of score decrease. To our knowledge, there have been no substantial attempts to build fully agentic systems for ITPs. The closest related effort is the early proof-of-concept by \citet{yang2023leandojo}, which explores the use of agents for proof generation in Lean. While this work illustrates the potential of agent-based approaches in this domain, it relies on minimal tooling and does not form a fully agentic system with autonomous planning and iterative execution.

As a user interface, we utilize CoqPilot to integrate into the common Rocq's programmer pipeline. CoqPilot is a VSCode%
\footnote{VSCode: \url{https://code.visualstudio.com}} plugin, facilitating access to Rocq generation methods for end-users. Currently, the RocqStar retrieval component is available in CoqPilot as a premise-ranking module. Integration of the full RocqStar agentic system as a proof generator is ongoing.

\section{Conclusion}\label{sec:conclusion}
We have presented a method to enhance retrieval-augmented generation in Rocq via leveraging neural premise selection using a self-attentive embedder model. We evaluated our proposed solution on a dataset of 300 Rocq theorems with two different generator models under the hood and showed a noticeable improvement of up to 28\% relative to the baseline. Our result suggests that proof-aware premise selection considerably improves generation quality, particularly for medium-difficulty theorems, where the gap between statement similarity and proof similarity becomes more significant. 

Our work pioneers the use of Agentic Systems applied to Formal Verification. We have implemented an advanced pipeline that includes rigorous planning via multi-agent debate, domain-specific tooling, and an adaptive executor–critic loop that iteratively refines proofs based on partial progress. We conclude that our RocqStar agent shows promising results, surpassing strong baselines and highlighting the applicability of agentic systems in the domain of theorem proving. The ablation study further demonstrates that both multi-agent planning and reflection are key to maintaining stable reasoning and achieving consistent improvements in the proof's success rate.

%%%%%%%%%%%%%%%%%%%%%%%%%%%%%%%%%%%%%%%%%%%%%%%%%%%%%%%%%%%%%%%%%%%%%%%%

\begin{acks}
We thank Ekaterina Verbitskaia, Ivan Kabashnyi, Maksim Rozenberg, and Pavel Guliaev for their valuable comments on this work. We also thank the Dynamic Program Analysis Research team at JetBrains (Nikita Dukin, Saga Rut Sunnevudóttir, and Egor Klimov) for their help in building tool execution distributions described in Appendix~\ref{appendix:executionpatters}. Finally, we thank the Koog team at JetBrains for developing Koog, an open-source framework for building AI agents with an idiomatic, type-safe Kotlin DSL.
\end{acks}

%%%%%%%%%%%%%%%%%%%%%%%%%%%%%%%%%%%%%%%%%%%%%%%%%%%%%%%%%%%%%%%%%%%%%%%%

\bibliographystyle{ACM-Reference-Format} 
\bibliography{rocqstar}

%%%%%%%%%%%%%%%%%%%%%%%%%%%%%%%%%%%%%%%%%%%%%%%%%%%%%%%%%%%%%%%%%%%%%%%%

\clearpage

\appendix

\section{Encoder Training Dataset}\label{appendix:encoderdataset}
One of the limitations of our BigRocq tool is that it cannot process theorems that contain so-called \emph{goal selectors}. The following example illustrates how they work. 
\begin{lstlisting}[language=coq, style=coqstyle]
Theorem test2nat1 : forall n : nat, n = 0 \/ n <> 0.
Proof.
    destruct n.
    - left; auto.
    - right; auto.
Qed. 
\end{lstlisting}
This example could be rewritten with the use of goal selectors to the following proof: 
\begin{lstlisting}[language=coq, style=coqstyle]
Theorem test2nat1 : forall n : nat, n = 0 \/ n <> 0.
Proof.
    intros n.
    destruct n.
    all: try (left; auto) || (right; auto).
Qed. 
\end{lstlisting}
Due to the limited information we get from the Coq-LSP, our heuristic algorithm of transforming the proof into a tree breaks down. We cannot augment such theorems. The authors of CoqGym~\cite{coqgym} also explicitly state that they do not handle theorems with goal selectors. They state that in their dataset, goal selectors occur in less than 1\% data. The situation has changed since the work was published; the feature is now used more often but is still relatively rare. Goal selectors are an issue to be solved, and we are working on a solution by extracting some additional information from Rocq's system through Coq-LSP\@.

The dataset is stored as a collection of JSON files and, due to its relatively small size, is stored within the repository, in the sub-directory with the model training code: \url{https://github.com/JetBrains-Research/rocqstar-rag/tree/main/proof-embeddings/data}.

Dataset is split into training, validation, and test sets with proportions of 70\%, 20\%, and 10\% respectively. Theorems from the same file do not appear in different sets. Parameters of building the dataset are listed in \Cref{tab:hp-model}. Pair of statements is considered as negative, if the \texttt{proof\_distance} between them is greater than 0.65, and positive if it is less than 0.3. If the distance is in range $[0.45, 0.65]$, with probability of 30\% it is also considered to be a negative pair (see \emph{hard negatives} in \Cref{sec:emb-modeling}).

\section{Encoder Details}\label{appendix:encodertraining}

\begin{table}[t]
  \centering
  \subfloat[Optimization hyperparameters\label{tab:hp-opt}]{
    \begin{minipage}[t]{0.5\textwidth}
      \centering
      \begin{tabular}{@{} ll @{}}
        \toprule
        \textbf{Parameter} & \textbf{Value} \\
        \midrule
        algorithm          & AdamW ($0.9,\;0.99,\;\mathrm{e}{-2}$) \\
        schedule           & linear warmup (10)               \\
        lr                 & $4\mathrm{e}{-6}$                \\
        batch size (stmts) & 16                               \\
        dropout            & 0.1                              \\
        \bottomrule
      \end{tabular}
    \end{minipage}
  }\hfill
  \subfloat[Model\&Dataset hyperparameters\label{tab:hp-model}]{
    \begin{minipage}[t]{0.45\textwidth}
      \centering
      \begin{tabular}{@{} ll @{}}
        \toprule
        \textbf{Parameter} & \textbf{Value} \\
        \midrule
        embedding dim              & 768     \\
        max sequence length        & 128     \\
        (positive, negative) threshold         & (0.3, 0.65) \\
        threshold hard neg.        & 0.45 \\
        hard negatives prob.       & 30\% \\
        \bottomrule
      \end{tabular}
    \end{minipage}
  }
  \vspace{0.2cm}
  \caption{Hyperparameters of the embedder training}\label{tab:hp-all}
  \vspace{-0.8cm}
\end{table}

The hyperparameters used for training the embedder model are listed in \Cref{tab:hp-all}. We have used \texttt{microsoft/codebert-base} as the base model and trained our embedder for 22000 steps with a batch size 16, since we use many negative samples in the loss. We applied a dropout of 0.1 on the last layer of the model; the embedding dimension is 768, and the maximum sequence length is 128. We use AdamW optimizer with a linear warmup schedule for 10\% of the training steps. 

\section{Agent toolset}\label{appendix:toolset}

Below are the tools that the agent uses to interact with Rocq's system. The session is a utility abstraction, mainly handled by our middleware server under the MCP\@. It manages sessions and creates a new one when the agent starts proving a new theorem. Sessions are introduced to speed up type-checking and reduce overhead. When the session is started, we create a file, copy all required theorem's context into it, type-check the context using Coq-LSP, and then start executing commands and continuously checking generated proofs in the context of this session. 

\begin{itemize}
  \item \textbf{\texttt{list\_coq\_files}}: Returns a list of all Coq files in the project.
  \item \textbf{\texttt{get\_theorem\_names}}: Retrieves theorem names available in the file, including the target theorem.
  \item \textbf{\texttt{get\_theorem\_names\_excl}}: Retrieves the available theorem names from a file with the target theorem excluded from the list.
  \item \textbf{\texttt{get\_current\_target\_state}}: Returns the proof stage for the target theorem in the current session.
  \item \textbf{\texttt{get\_theorem\_with\_proof}}: Given the theorem's name, returns the theorem with its proof.
  \item \textbf{\texttt{check\_proof}}: Validates a proof (or a part of a proof) in the context of a session and returns either of the following: 
  \begin{itemize}
    \item[(i)] That there are no more goals to prove
    \item[(ii)] Provided proof produces no errors, but the goal is not fully solved. Returns: updated goal state
    \item[(iii)] The current goal is solved, but there are more goals at other depth levels. Returns: first unsolved goal at the closest depth level
    \item[(iv)] Provided proof produces errors. Returns: error message
  \end{itemize}
  \item \textbf{\texttt{get\_similar\_proofs}}: Given theorem goal/statement as a string, it uses RocqStar ranker to retrieve theorems that are similar to the input statement and returns 15 most similar ones. 
  \item \textbf{\texttt{about\_term}}: Uses \texttt{About} Rocq's Command in the current session. Accepts the term name as an argument. Outputs the term's definition and a short description from the Coq-LSP\@.
  \item \textbf{\texttt{search\_pattern}}: Uses Rocq's \texttt{Search ?exp} to search for a pattern in the current session's file. An example of a valid command: \texttt{Search (?a + ?b = ?b + ?a)}. It could be useful for finding lemmas that could be used in the proof.
  \item \textbf{\texttt{print\_term}}: Prints a term in the current session file. Uses Rocq's \texttt{Print} command. Accepts the term name as an argument. Outputs the term's definition.
  \item \textbf{\texttt{check\_term}}: Checks a term in the current session's file. Uses Rocq's \texttt{Check} command. Similarly to \texttt{print\_term} or \texttt{about\_term}, but outputs only the type of the term. In the case of a theorem, it outputs its statement.
\end{itemize}

\section{IMM Evaluation Dataset}\label{appendix:evaldataset}
The collected \emph{IMM-300} dataset from the CoqPilot~\cite{coqpilot} work includes only theorems with proofs of length no more than 20. For that reason the bucket with the most difficult theorems is labeled 9---20 tactics. This decision has been made, reflecting CoqPilot's original focus on subgoals and shorter lemmas. Theorems of length no more than 20 tactics account for 83\% of all proofs in the IMM project. As we take the same dataset, it possesses the same limitations. Therefore, we have not evaluated our solution on theorems, for which the reference proof contains more than 20 tactics. However, such theorems are quite rare.

The exact list of theorems used in each group could be found in the repository of the CoqPilot project: \url{https://github.com/JetBrains-Research/coqpilot/blob/main/etc/docs/benchmark/}.

A common problem with testing pipelines that include general-purpose LLM providers, such as OpenAI, is data contamination. We are aware, that the model could have possibly seen the human-written proofs, as the IMM project has been publicly available since a while. However, firstly, the model is provided with neither the theorem name for which it is generating the proof, nor the proof goal exactly as it appeared in the original file. As we treat them as proof states, rather than theorems, an LLM receives it in an equivalent, but slightly modified way. Secondly, as many of our experiments have shown, various quality of premise selection drasticly changes the behavior of the model. That hints that the model is not able to memorize all theorems and proofs. Lastly, data contamination issue was one of the things we had in mind, while developing BigRocq. One could pass a Rocq project into BigRocq as input, and for each theorem retrieve the sub-state, that is achieved after $k$ steps. On an example of $k = 2$, the following theorem: 
\begin{lstlisting}[language=coq, style=coqstyle]
Lemma eq_trans (A : Type) : forall (x y z : A), x = y -> y = z -> x = z.
Proof.
    intros x y z Hxy Hyz.
    rewrite Hxy. (* State: (A : Type) (x y z: A) (Hxy: x = y) (Hyz: y = z) : y = z *)
    rewrite Hyz.
    reflexivity.
Qed.
\end{lstlisting}
Could be automatically tranformed into the following one: 
\begin{lstlisting}[language=coq, style=coqstyle]
Lemma eq_trans_modified (A : Type) (x y z: A) (Hxy: x = y) (Hyz: y = z) : y = z.
Proof.
    rewrite Hyz.
    reflexivity.
Qed.
\end{lstlisting}
The higher $k$ is chosen, the smaller would be the chances of data leakage, as the produced sub-state gets further and further from the original theorem.

\subsection{Visualizing RocqStar vs. Baseline Premise Selection}

Here we try to illustrate the difference between different rankers and show an example of a theorem from IMM project, where our ranker outperforms the baseline. Figure~\ref{fig:rankers-viz} presents such an example. 

\begin{figure}[ht]
    \centering
    \begin{minipage}[t]{.495\linewidth}
\noindent\fcolorbox{black}{filegray}{%
\parbox{\dimexpr\linewidth-2\fboxsep-2\fboxrule-6pt\relax}{%
   \ttfamily\scriptsize
   \raggedright{}
   \textcolor{blue}{Lemma}\ ext\_sb\_trans : transitive \highlight[red]{exts\_sb}.\\
   Proof using.\\
   \quad \highlight{unfold ext\_sb; red; ins.}\\
   \quad \highlight{destruct x},y,z; \highlight{ins; desf; splits;} eauto.\\
   \quad by rewrite H2.\\
   \textcolor{blue}{Qed}.\\[1ex]
 }
}
    \end{minipage}
    \hfill
    \begin{minipage}[t]{.495\linewidth}
\noindent\fcolorbox{black}{filegray}{%
\parbox{\dimexpr\linewidth-2\fboxsep-2\fboxrule-1pt\relax}{%
   \ttfamily\scriptsize
   \raggedright{}
   \textcolor{blue}{Lemma}\ ext\_sb\_irr : irreflexive \highlight[red]{exts\_sb}.\\
   Proof using.\\
   \quad \highlight{unfold ext\_sb; red; ins.}\\
   \quad \highlight{destruct x}; \highlight{ins; desf; splits;} firstorder.\\
   \quad lia.\\
   \textcolor{blue}{Qed}.\\[1ex]
 }
}
    \end{minipage}
    \caption{Theorems with dissimilar statements and similar proofs}\label{fig:rankers-viz}
\end{figure}

If we measure the distance between theorems from Figure~\ref{fig:rankers-viz} using the conventional Jaccard distance, which is used by default in CoqPilot, we get $0.67$:
{\footnotesize
\begin{align*}
    \operatorname{Jaccard\_dist}(t1, t2) &= 1 - \frac{\left|\{\thrtoken{transitive}, \thrtoken{ext\_sb}\} \cap \{\thrtoken{irreflexive}, \thrtoken{ext\_sb}\}\right|}{\left|\{\thrtoken{transitive}, \thrtoken{ext\_sb}\} \cup \{\thrtoken{irreflexive}, \thrtoken{ext\_sb}\}\right|} \\ 
    &= 1 - \frac{1}{3} = 0.67
\end{align*}}
Jaccard ranker focuses only on statement similarity, which in this case is relatively small, the only similar parts are highlighted with \highlight[red]{red}. Jaccard would probably not select theorem \texttt{ext\_sb\_irr} as a premise for theorem \texttt{ext\_sb\_trans}; however, they have similar proofs and one could help the model to generate the proof for the other. Similar parts of the proofs are highlighted with \highlight{yellow}. If we measure the distance between these theorems using the \texttt{proof\_similarity} metric we define, we get $0.32$, and our trained model yields $0.28$. When using our ranker, it is probable that one theorem would be selected as a premise for the other.
\begin{align*}
    \operatorname{proof\_sim}(t1, t2) &= 0.32 \\ 
    \operatorname{embedder\_pred}(t1, t2) &= 0.28
\end{align*}

\section{Experiments compute resources}\label{appendix:expresources}
Our experiments in the evaluation part mainly use cloud LLM providers and therefore require minimum compute, but comprehensive financial resources.

\textbf{Embedder training}\; During our final training run of the embedding model for RocqStar ranker, the model consumed roughly 43 GB of GPU-process memory and only about 6\% of the host's RAM\@. Over 31 hours on a single NVIDIA H200 accelerator (with 16 CPU cores and 200 GB of system memory), disk usage grew steadily from 28 GB to 76 GB as checkpoints and logs accumulated. GPU utilization stabilized above 90\% shortly after the warmup phase and remained near saturation for the remainder of training. To sum up, our setup runs comfortably on a single GPU node with modest additional CPU and memory overhead.

\textbf{Embedder model evaluation}\; Experiments were conducted on a single MacBook Pro with an M1 chip. The only computationally expensive part of the experiments (for the local machine) is launching multiple Coq-LSP servers at once (CoqPilot benchmark does that to optimize the time of the experiments and accelerate type-checking). As we use a middleware service over LLM APIs, our financial estimations might not be accurate. However, we roughly estimate 12 generation attempts per theorem with seven contextual theorems at 12 cents per theorem for Claude 3.5 and 7 cents for GPT-4o. Running the experiment on 300 theorems and repeating it three times amounts to a total of 171 US Dollars. Performing the same experiment for three different retrieval engines brings the overall cost to approximately 513 US Dollars.

\textbf{Agent evaluation}\; In the case of the agent evaluation, we ran the experiment only once and did not provide the confidence intervals due to financial limitations. We ran our agent on the IMM-300 dataset, and afterward, we compared five versions of the agent on the IMM-50 dataset in our \Cref{sec:ablationstudy}. That results in 550 attempts to prove different theorems. We estimate a single attempt at 1.3 US dollars. Therefore, we estimate the cost of the evaluation of the agent to be 715 US dollars.

\section{Plan refinement via Multi-Agent Debate}\label{appendix:planningmad}

We compare two proof plans: one generated by a single model call and one produced via multi-agent debate. The target theorem is \texttt{sub\_IPROP} from \texttt{src/imm/SubExecution.v}, which states:
\begin{lstlisting}[language=coq, style=coqstyle]
Lemma sub_IPROP : subseteq (IPROP G') (IPROP G).
\end{lstlisting}

The following plan was produced by a single planning call:

\begin{tcolorbox}[
  colback=filegray,
  left=1em, right=1em,
  top=1em, bottom=1em,
  boxrule=0pt,
  arc=1mm
]
\begin{ttfamily}
Let me outline a proof strategy for showing that \texttt{IPROP G'} is a subset of \texttt{IPROP G}.

First, I want to understand the context better. \highlight[red]{The lemma involves two grammars \texttt{G} and \texttt{G'}, and we need to show that the improper symbols of \texttt{G'} are contained within the improper symbols of \texttt{G}.} Let me check the definitions and related theorems.

\medskip
\texttt{print\_term: "IPROP"}

Understanding the definition of \texttt{IPROP} will help us formulate the proof. \highlight[red]{This looks like a lemma about grammar properties, where we need to show that if a symbol is improper in \texttt{G'}, it must also be improper in \texttt{G}.}

My proposed strategy:
\begin{enumerate}
  \item Use the definition of subset (\(\subseteq\)) to unfold the goal.
  \item For any symbol \texttt{X} in \texttt{IPROP G'}, show it’s also in \texttt{IPROP G}.
  \item This will likely require using the definitions of improper symbols and their relationship between grammars \texttt{G} and \texttt{G'}.
\end{enumerate}

I expect we’ll need to:
\begin{itemize}
  \item Start with \texttt{intros X H} to get a symbol \texttt{X} and hypothesis that it’s in \texttt{IPROP G'}.
  \item Use the definition of improper symbols to understand why \texttt{X} is improper in \texttt{G'}.
  \item Show that these same conditions must hold in \texttt{G}.
\end{itemize}

Before proceeding with the actual proof steps, I’d like to see if there are any similar theorems or relevant properties about \texttt{IPROP} that might help us.

\medskip
\texttt{search\_pattern: "IPROP"}
\end{ttfamily}
\end{tcolorbox}

The proposed proof strategy is quite vague, too wordy, and imprecise. With this plan, the executor agent started by fetching similar theorems and then got stuck in a cycle of ad-hoc attempts to check incorrect proofs.
In contrast, the winning plan was precise, concise and clear. It had a top-level structure of the proof incorporated into the plan and made a winning solution to use the \texttt{search} command, which helped to finish the proof: 
\begin{lstlisting}[language=coq, style=coqstyle]
Proof.
    unfold IPROP.
    rewrite sub_W. rewrite sub_is_ta_propagate_to_G.
    done.
Qed.
\end{lstlisting}

Below, as a comparisson, is the plan, yielded by the multi-agent debate: 
\begin{tcolorbox}[
  colback=filegray,
  left=1em, right=1em,
  top=1em, bottom=1em,
  boxrule=0pt,
  arc=1mm
]
\begin{ttfamily}
  The winning proof plan follows these steps:

  \begin{enumerate}
    \item \textbf{Initial Analysis:}
      \begin{itemize}
        \item Use \texttt{rewrite /IPROP /subseteq} to fully understand the expanded goal
        \item \highlight[green]{Check \texttt{G} and \texttt{G'} relationship via \texttt{about\_term} and \texttt{search\_pattern}}
        \item Look for helper lemmas about \texttt{IPROP} transformations
      \end{itemize}
  
    \item \highlight[green]{\textbf{Proof Structure:}}
  
    \medskip
    \begin{ttfamily}
    Proof.\\
    rewrite /IPROP /subseteq.\\
    (* See expanded goal structure *)\\
    intros x Hx.\\
    (* Analyze Hx’s type carefully *)\\
    (* Use discovered helper lemmas *)
    \end{ttfamily}
    \medskip
  
    \item \textbf{Reasoning Steps:}
      \begin{itemize}
        \item Work with explicit set inclusion based on \texttt{subseteq} definition
        \item Use any discovered lemmas about \texttt{G/G'} relationship
        \item Preserve flexibility to adapt based on actual definitions
      \end{itemize}
\end{enumerate}
\medskip
\textbf{B’s plan wins because it:}
\begin{itemize}
  \item \highlight[green]{Prioritizes understanding definitions before tactical decisions}
  \item Makes systematic use of available tools
  \item Avoids assumptions about existential structure
  \item Maintains flexibility while having concrete next steps
\end{itemize}
\end{ttfamily}
\end{tcolorbox}

\section{Agent Execution Patterns}\label{appendix:executionpatters}

\begin{figure}[t]
  \centering
  \begin{minipage}{0.48\textwidth}
    \centering
    \includegraphics[width=\linewidth]{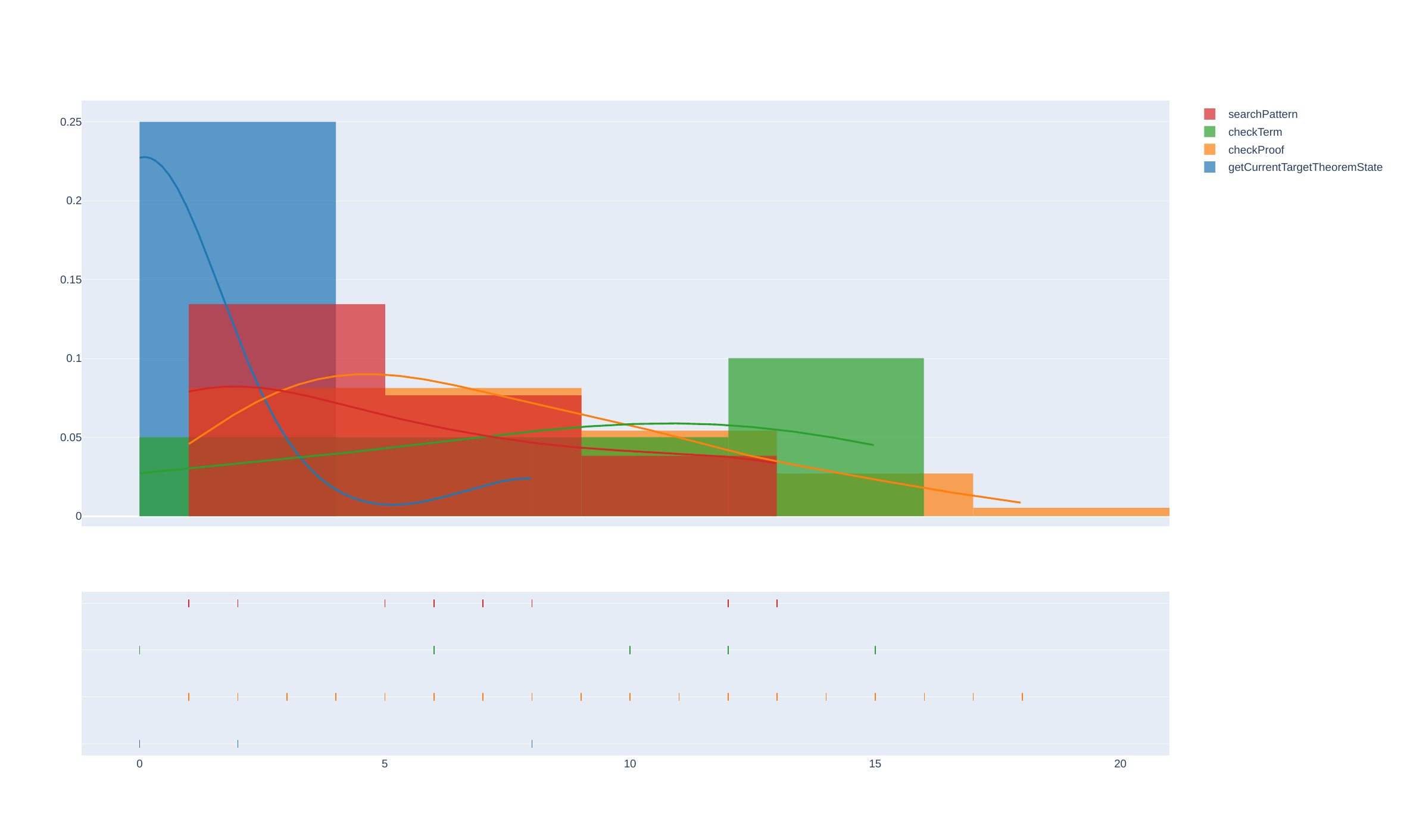}
    \caption*{Average successful run}
  \end{minipage}\hfill
  \begin{minipage}{0.48\textwidth}
    \centering
    \includegraphics[width=\linewidth]{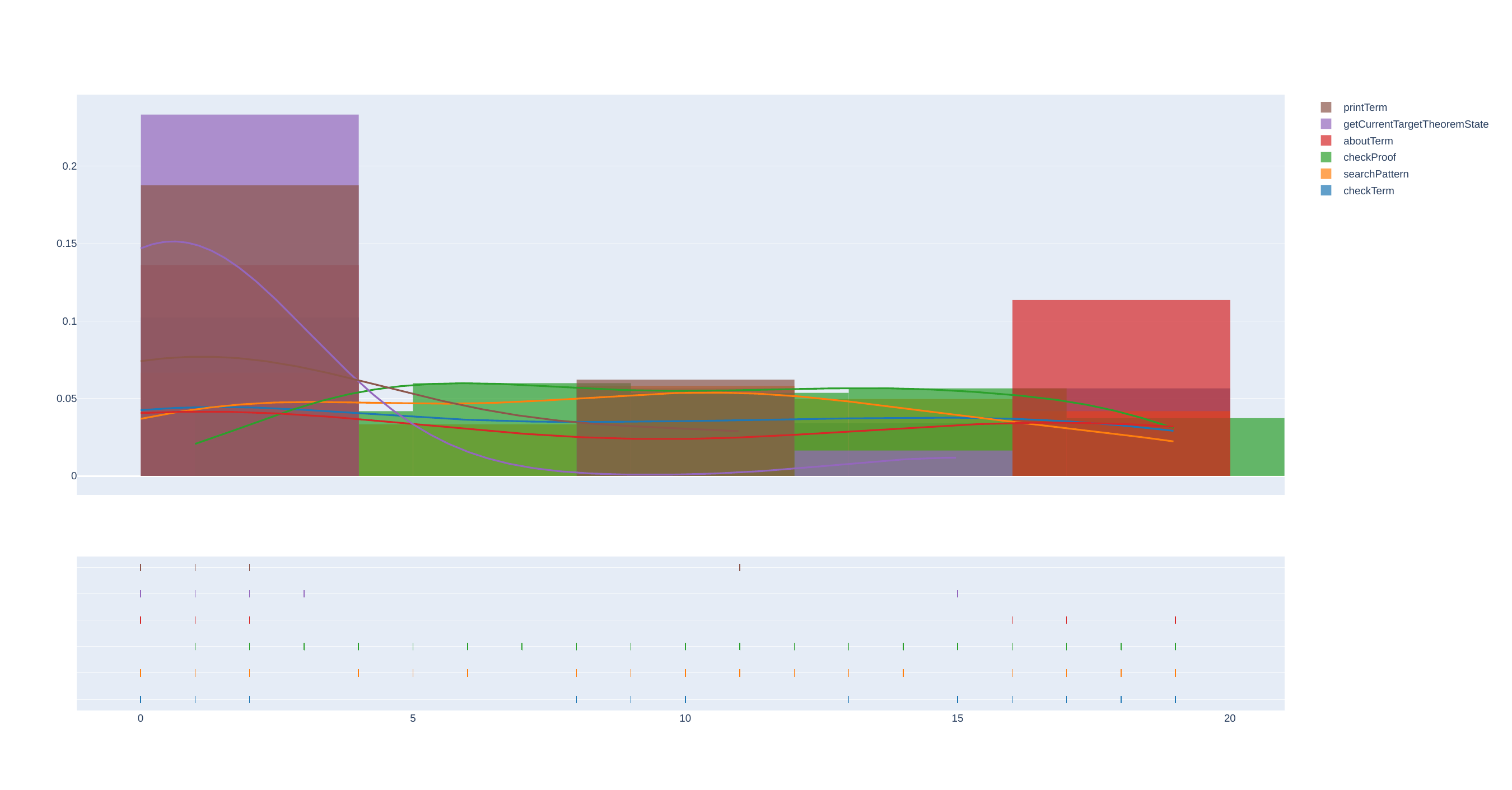}
    \caption*{Average failing run}
  \end{minipage}
  \caption{Distributions of successful and failing runs for the \texttt{no\_sb\_cr\_to\_init} theorem}
  \label{fig:proofpairs}
\end{figure}

We analyze the agent’s behavior to identify patterns that distinguish successful and failed proof attempts.
Specifically, we selected the theorem \texttt{no\_sb\_cr\_to\_init} from group B of the IMM-50 dataset, which the agent sometimes proved and sometimes failed to prove.
We collected traces from ten successful and seventeen failing executions and plotted the distributions of tool calls averaged over all runs in each scenario (Fig.~\ref{fig:proofpairs}). 

Although the experiment was relatively small, it already provides several observations about the agent’s behavior. We have not yet identified clear action points that could directly improve the agent, but we plan to conduct a larger-scale study and dedicate more effort to analyzing execution trajectories in the future.
Notably, only a subset of available tools is actively used during proof attempts, especially in successful runs, indicating that the agent has not yet learned to leverage the full range of available functionality, although this behavior partly reflects the specifics of the analyzed theorem.
In failing runs, we observe a more uniform distribution of tool calls (particularly \texttt{aboutTerm}, \texttt{printTerm}, and \texttt{searchPattern}), without a pronounced early activity peak. This behavior suggests that the agent tends to wander through the search space without a clear strategy. Combined with the findings from \Cref{sec:ablationstudy}, which highlight the importance of effective planning, this supports the view that a well-defined plan, one that explicitly incorporates context exploration and the retrieval of relevant information before trying to build the proof, is essential for stable performance. We consider introducing a dedicated, more manual phase for contextual exploration in future iterations of the agent.

%%%%%%%%%%%%%%%%%%%%%%%%%%%%%%%%%%%%%%%%%%%%%%%%%%%%%%%%%%%%%%%%%%%%%%%%

\end{document}